# The Advancement of Personalized Learning Potentially Accelerated by Generative AI

Yuang Wei[1, 2, 3, 5], Yuan-Hao Jiang[1, 2, 4, 5, *], Jiayi Liu[1, 5], Changyong Qi[1, 2, 5], Linzhao Jia[1, 2, 5], Rui Jia[1, 2, 5]

[1] Lab of Artificial Intelligence for Education, East China Normal University, Shanghai, 200062, China
[2] School of Computer Science and Technology, East China Normal University, Shanghai, 200062, China
[3] School of Computing, National University of Singapore, 119077, Singapore
[4] Graduate School, Shanghai Jiao Tong University, Shanghai, 200030, China
[5] Shanghai Institute of Artificial Intelligence for Education, East China Normal University, Shanghai, 200062, China
[*] Corresponding Author: yuanhao.cs.edu@gmail.com (Yuan-Hao Jiang)

**Abstract:** The rapid development of Generative AI (GAI) has sparked revolutionary changes across various aspects of education. To explore GAI's extensive impact on personalized learning, this study investigates its potential to enhance various facets of personalized learning through a thorough analysis of existing research. The research comprehensively examines GAI's influence on personalized learning by analyzing its application across different methodologies and contexts, including learning strategies, paths, materials, environments, and specific analyses within the teaching and learning processes. Through this in-depth investigation, we find that GAI demonstrates exceptional capabilities in providing adaptive learning experiences tailored to individual preferences and needs. Utilizing different forms of GAI across various subjects yields superior learning outcomes. The article concludes by summarizing scenarios where GAI is applicable in educational processes and discussing strategies for leveraging GAI to enhance personalized learning, aiming to guide educators and learners in effectively utilizing GAI to achieve superior learning objectives.

## 1 Introduction

Two millennia ago, Confucius introduced the revolutionary concept of "teaching students according to their aptitude," (Tan, 2017) laying the groundwork for individualized teaching. This transformative idea re-emerged in John Dewey's groundbreaking research and practice, advocating for the recognition of individual differences and charting a course for flexible, personalized education in his visionary "School of Tomorrow." This philosophy propelled the early 20th century's new education movement, leading to significant shifts from traditional, one-size-fits-all teaching approaches to methods that emphasize personalized learning strategies, cooperative learning, and activities tailored to students' interests and needs. Despite these advances, the large-scale implementation of personalized learning has historically faced formidable challenges due to technological limitations (X. Li et al., 2025; Wang, Tu, et al., 2021; Wang, Zheng, et al., 2021).

However, traditional ITSs often demand extensive expert input for content and course structure creation, limiting their scalability and personalization. To address these limitations, researchers have delved into the 'inner loop' of ITS, focusing on specific instructional guidance like feedback and evaluations (Koedinger et al., 2013). Despite advancements in big data, computing, and algorithms, constructing efficient inner loops remains costly and inefficient (Afzal et al., 2019; Anderson & Skwarecki, 1986). Conversely, the 'outer loop' is more adaptable, encompassing learner profile analysis and task decision-making. Yet, long-standing technical challenges have hindered accurate analysis and decision-making in this domain.

This is where Generative AI (GAI) comes in. The advent of GAI, particularly large language models like ChatGPT, has dramatically accelerated the progress of personalized learning (Jiang, Li, Wei, et al., 2024; Jiang, Shi, et al., 2024). But how exactly does GAI propel personalized learning forward at such a remarkable pace? There is still no comprehensive summary of its contributions. So, this paper aims to bridge that gap by thoroughly exploring the origins, various models, and applications of GAI in personalized learning. We categorize these applications into Learning Strategies, Paths, Teaching Materials, and Learning Environments (Fariani et al., 2023). Furthermore, we examine the transformative impact of GAI on teaching and learning practices.

## 2 Assisting in Generating Learning Strategies

Since 1987, education research has focused on developing effective learning strategies, ranging from historical initiatives like the Latin School Movements (Burleigh, 1973) to the extensive Head Start Program (Sternberg, 1982). These strategies are behaviors designed to influence information processing (Mayer, 1988), such as highlighting key points or summarizing lectures (Mayer, 1987). Among these, question prompts are crucial for guiding students through problem-solving, utilizing a scaffolding approach particularly effective in math education (Boston & Candela, 2018; Kim et al., 2018; Wood, 1994). The advent of GAI has revolutionized methods like Socratic questioning (Fig.1), which encourages deep thinking by posing targeted questions (Quintana et al., 2018). GAI tools can mimic this by dynamically generating questions based on students' interactions, fostering deeper analysis and active engagement in learning (R. Li, Jiang, et al., 2024; R. Li, Wang, et al., 2024). Recent studies have applied large language models for generating Socratic questions in math, focusing on goal-driven queries that enhance comprehension and critical thinking (Anghileri, 2006; Radford et al., 2019b; Wood, 1994).

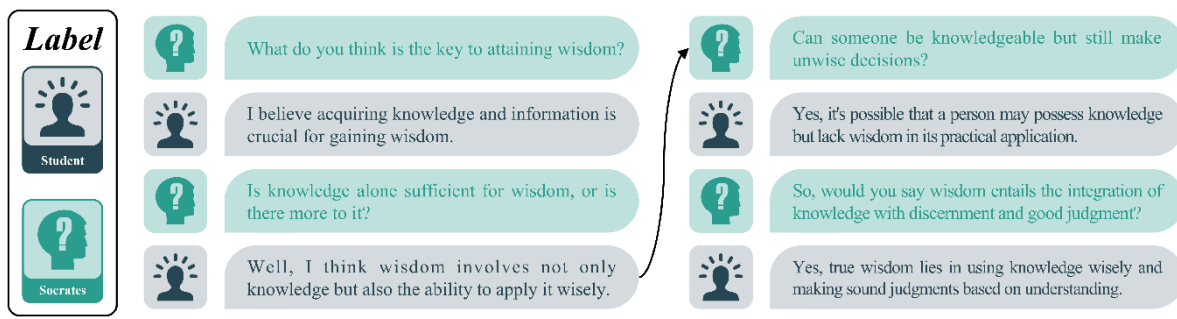

**Fig. 1** Socratic Q&A generated by ChatGPT

GAI enhances programming education by offering personalized exercises and tasks tailored to students' coding skills. It combines theory with practice, supporting mastery of languages, algorithms, and problem-solving. As a programming tutor, GAI provides real-time feedback, code analysis, and optimization suggestions (Zhou et al., 2024). Specifically, the strategies for programming education are consistent with those of other subjects. Programming courses and assignments typically progress in complexity, gradually introducing new concepts while ensuring that students are not overwhelmed, thus avoiding cognitive overload (Duran et al., 2022) and fostering students' proximal development (Vygotsky & Cole, 1978). The design of programming courses and assignments often aligns with the concept of deliberate practice (Ericsson et al., 1993), facilitating mastery through consistent practice(Edwards et al., 2020). Apart from creating programming assignments, teachers frequently provide feedback to students on their coursework (Ala-Mutka, 2005; Ihantola et al., 2010; Paiva et al., 2022). In recent research, used Codex to formulate more personalized programming learning strategies (Sarsa et al., 2022). Similar to the GPT (Brown et al., 2020)model, Codex (Chen et al., 2021)is a large language model developed by OpenAI with a specific focus on comprehending and generating programming languages. Sarsa and colleagues (Sarsa et al., 2022)investigated novel methods to create programming exercises that enhance the quality of generated content by introducing concepts in a structured manner. They also incorporated step-by-step explanations, as it aligns with the multi-structural levels of the SOLO taxonomy and is often generated by students when prompted to explain code (Lister et al., 2006). This learning strategy makes it easier for students to achieve their learning objectives. This structure is shown in the Fig. 2.

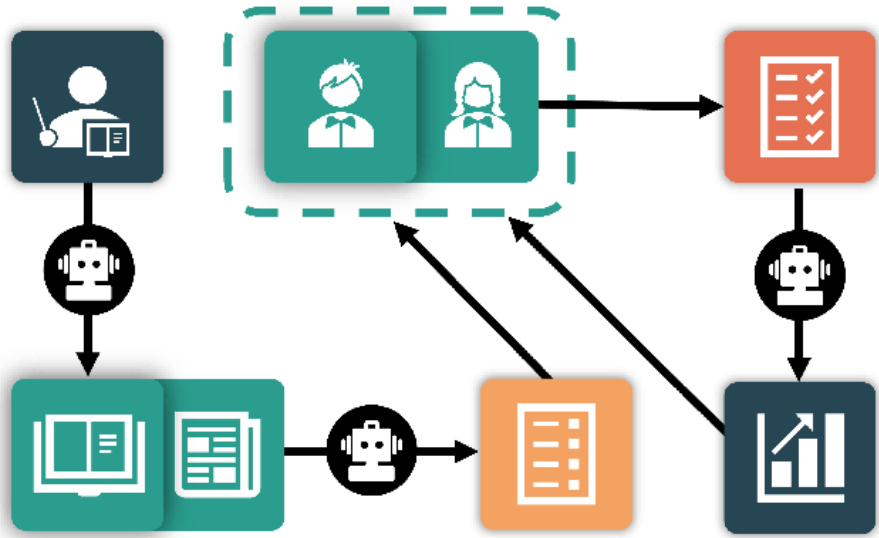

**Fig. 2** The Lifecycle of Programming Exercises: (1) Teachers create programming exercises and learning materials. (2) Students study the materials and practice the exercises. (3) Students receive feedback

## 3 Planning Personalized Learning Path

Personalized learning paths implement strategies tailored to individual learners' backgrounds and goals, employing technology to redefine educational approaches (Nabizadeh et al., 2020; Wei et al., 2024). E-learning, notably through Intelligent Tutoring Systems (ITS), has become integral, offering resources more flexibly and cost-effectively than traditional settings (Gilbert, 2015; Nabizadeh et al., 2017; Thakkar & Joshi, 2015). These systems facilitate anytime learning and improve collaboration (Dargham et al., 2012; Thakkar & Joshi, 2015). Recent advancements include the use of Generative AI (GAI) in ITS for generating customized educational prompts. This technique enhances learning efficiency by integrating keyword recognition, sentence analysis, and prompt generation using language models, further supplemented with Wikipedia-based explanations to aid comprehension (Kochmar et al., 2022). This approach has been empirically validated to boost student performance significantly.

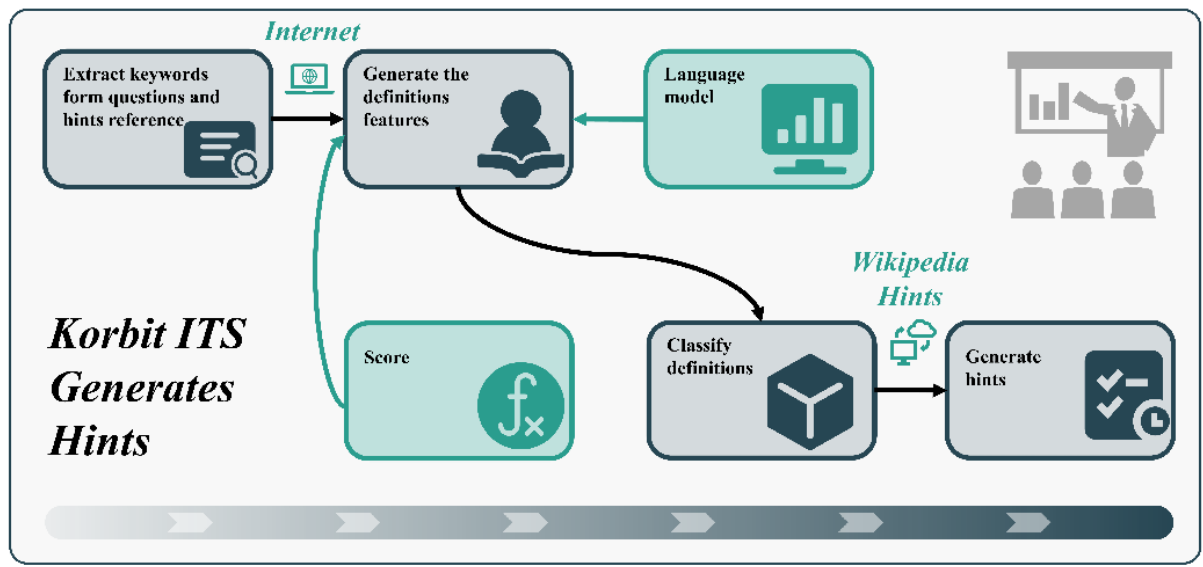

**Fig. 3** Korbit ITS generates hints

On the other hand, research has found that structuring learning experiences in the form of interactive "narratives" or "stories" told by the learning environment can enhance learner engagement(Marsh et al., 2011). Making the learning approach more interactive and aligning learning path presentation with typical narrative structures can expedite students' achievement of learning goals (Diwan et al., 2023). In Fig. 4, an AI-based approach was proposed to automatically generate learning content and integrate it into the appropriate positions within the learning path (Diwan et al., 2023). In other words, the interactive narrative sequences that conform to general narrative expectations are constructed by intertwining the learning path with the automatically generated learning content referred to as narrative fragments. The method proposed in this study constructs this interactive narrative in a domain-agnostic manner (Diwan et al., 2023). It can generate narrative fragments for learning resources in any format, unlike many AI-based learning environments that accept only specific formats of learning sources. Though narrative fragments can be created manually, they would cost a large amount of time and effort. Additionally, AI-based learning content generation methods can achieve adaptive learning by dynamically generating content based on learner-driven variations. Ultimately, the model built using GPT-2 (Radford et al., 2019a) has shown excellent results in empirical research.

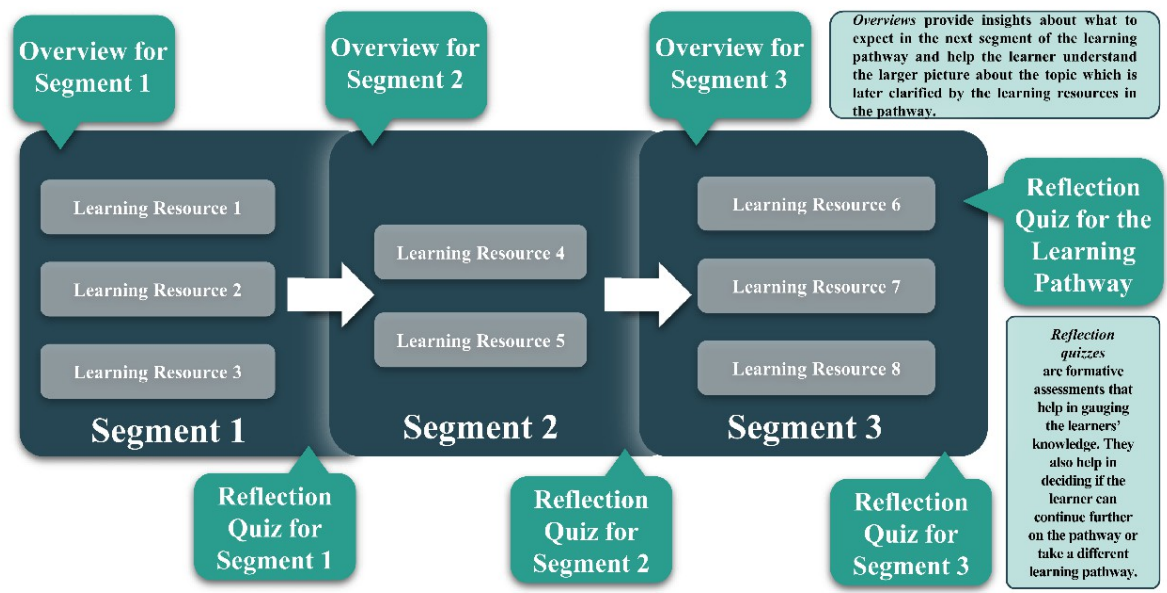

**Fig. 4** Learning Pathway showing different segments and addition of narrative fragments

## 4 How does Generative AI transform teaching and learning

**4.1 Assistance for Teaching**

GAI can support teaching by reducing teacher workload and enhancing efficiency. Effective integration requires understanding both AI capabilities and teachers' roles. This analysis examines GAI's potential to aid teachers in planning, implementation, and assessment, alongside the roles teachers play in these processes.

**4.1.1 Planning, implementation and assessment**

GAI integration in educational planning aids teachers in creating instructional materials and fostering interactive teaching practices. GAI generates personalized content and resources, such as GPT-4's learning objectives for AI courses (Sridhar et al., 2023) and a Human-NLP system for reading quiz questions (Lu et al., 2023). Applications include adaptive courseware (Schroeder et al., 2022), personalized practice problems(Jiang, 2024), and ChatGPT-assisted lesson planning (van den Berg & du Plessis, 2023). ChatGPT helps teachers with instructional goals and strategies, enhancing classroom discussions with examples and scenarios. Markel et al. developed 'GPTeach' for teacher training, allowing practice with simulated students (Markel et al., 2023). GAI enables teachers to craft high-quality pedagogical plans and curate resources, dynamically refining plans based on simulated student responses.

During implementation, GAI assists teachers in providing immediate, tailored feedback. Large language models offer feedback on writing tasks (Kasneci et al., 2023), and in programming education, they provide feedback on syntax errors with a validation mechanism (Phung et al., 2023). Machine-learning approaches in ITS automate feedback generation, improving students' success rates with personalized hints and explanations (Kochmar et al., 2022). Teachers supplement AI feedback by providing sub-step feedback, refining system feedback, and offering emotional support. Emotional support from teachers fills gaps in AI's ability to provide a human touch, facilitating social interaction and contextualizing learning experiences (Chan & Tsi, 2023).

GAI aids in automating evaluations and providing insights for assessment improvement (Hu et al., 2024). GPT-3 has been used for automated essay scoring with reliable accuracy on TOEFL11 essays (Mizumoto & Eguchi, 2023), and to simplify math word problems, improving readability (Patel et al., 2023). Assessments benchmarked against GPT identified strengths and weaknesses in engineering education, offering recommendations for future assessment design (Nikolic et al., 2023). Teachers must supervise GAI output and design comprehensive assessments, verifying AI-generated evaluations (Celik et al., 2022) and monitoring GAI performance (Baidoo-Anu & Ansah, 2023). Diversified assessment methods, including formative and summative assessments, group projects, hands-on activities, and oral presentations, are necessary for comprehensive evaluation of students' abilities (Tlili et al., 2023). Collaborative research is needed to further explore the roles of teachers and GAI in assessment.

**4.1.2 Complementary relationships between teachers and GAI in education**

The relationship between GAI and teachers in adaptive learning environments emphasizes hybrid intelligence, shifting from AI replacing teachers to augmenting their capabilities. The most effective adaptive learning combines AI and human facilitators (Holstein et al., 2020; Jeon & Lee, 2023; Molenaar, 2022). Holstein et al. identify four key areas where AI and teachers can enhance adaptability: goal, perceptual, action, and decision augmentation (Holstein et al., 2020). Jeon and Lee explore role division between teachers and GPT to maximize teaching effectiveness, showing potential for collaboration in curriculum planning, teaching implementation, and evaluation (Jeon & Lee, 2023). Bai et al. note the evolving role of teachers in AIED, focusing on guiding learning, assessing performance, and offering personalized support (Bai et al., 2022). AIED tools enable teachers to better understand students' needs and progress by analyzing data, leading to more personalized teaching plans. These tools also assist in classroom management and supervision, allowing real-time monitoring of student engagement and automating grading to reduce workload and improve efficiency.

However, caution is needed to avoid over-reliance on algorithms, which might overlook human factors or misuse data. Teachers must ensure AIED tools enhance their role in promoting student learning and development. Despite progress, empirical studies on the collaboration between teachers and GAI in adaptive learning are limited, requiring further research to optimize this synergy.

**4.2 Assistance for Learning**

From learners' perspective, GAI holds the potential to offer multifaceted assistance to the learning process. The diverse range of applications provided by GAI can be roughly categorized into three kinds according to the type of content they provide, including direct solutions, hints and instructional cues, and heuristic dialogues.

**4.2.1 Provide learners with direct solutions**

Generative AI assists learners by providing solutions to various problems, including math, coding, and essay writing. Wardat et al. found that ChatGPT proficiently executes mathematical operations, manipulates algebraic expressions, and solves complex calculus problems (Wardat et al., 2023). Poesia et al. enhanced code generation reliability using large pre-trained language models. This capability allows learners to quickly access learning resources and potential solutions to their problems (Poesia et al., 2022). A study (Nurmayanti & Suryadi, 2023) explored Quillbot's impact on students' writing skills, finding it effectively helped students preprocess text and reduce plagiarism. The study, conducted in three stages—planning, implementation, and evaluation—showed improved paraphrasing skills among English education majors. However, the accuracy of GAI-generated solutions can vary with problem complexity and context, necessitating double-checking by learners (Poesia et al., 2022; Wardat et al., 2023). As GAI advances, more reliable solutions are expected to enhance learner support across various scenarios.

**4.2.2 Offer instructional cues and hints to learners**

Another approach for GAI to assist learners is by offering instructional cues and hints. Constructivism emphasizes learners' active role in constructing knowledge through hands-on experience or problem-solving. Providing hints and scaffoldings can aid learners without harming their initiative (Jiang, Li, Zhou, et al., 2024). Pankiewicz and Baker (Pankiewicz & Baker, 2023) used GPT-3.5 to generate personalized hints for programming assignments, which were positively received by students and improved their task-solving abilities. An experiment with a Chatbot-Assisted Classroom Debate (CaIcD) involved students using a chatbot named Argumate to support their viewpoints and present opposing perspectives. This helped students anticipate counterarguments and prepare effective rebuttals, leading to more organized and substantial arguments and higher motivation (Guo et al., 2023). However, students showed a lower success rate initially when switched to more complex tasks without GPT-generated hints, indicating potential reliance on GAI (Pankiewicz & Baker, 2023). A study (Jin et al., 2023) found AI applications beneficial for supporting metacognitive, cognitive, and behavioral regulation aspects in online education but less effective for motivational regulation. Abdelghani et al. (Abdelghani et al., 2023) used GPT-3 to generate pedagogical cues to stimulate children's curiosity, enhancing their question-asking abilities. Other examples of GAI-generated scaffoldings include code explanations (Sarsa et al., 2022) and reading quizzes (Lu et al., 2023). Learners can leverage GAI-provided cues and scaffoldings to improve their learning experience.

**4.2.3 Facilitate Heuristic Dialogues for Learning**

Generative AI (GAI) can utilize the "Socratic Method" to enhance learning through discussions (Qi et al., 2024), Q&A sessions, and debates, promoting critical thinking by guiding students in independent analysis and understanding of complex concepts. Recent studies have used GAI to implement Socratic questioning strategies: Shridhar et al. demonstrated that large language models (LMs) could generate sequential questions to improve performance in solving simple math problems, though efficiency declined with complexity. Another study tested GPT-based models in Socratic debugging for novice programmers, finding that human experts still outperform these models (Al-Hossami et al., 2023)., with GPT sometimes generating repetitive or irrelevant questions. Future research should explore further integrations of GAI with the Socratic method across diverse learning contexts.

**4.3 Ethical Issues of Generative AI in Education**

Despite the significant potential of GAI to promote educational advancement, its deployment within educational settings introduces multiple ethical challenges that necessitate careful consideration to ensure equitable access to technology and to address biases inherent in AI models. A principal ethical concern involves guaranteeing fair access to AI-enhanced educational tools, which may be impeded by socioeconomic disparities. Initiatives such as providing subsidies for technology access in economically disadvantaged areas or offering additional discounts in less developed regions are critical measures to alleviate these challenges. Additionally, tackling AI biases necessitates the execution of extensive testing phases aimed at identifying and amending any biased AI responses prior to their impact on students. It is equally imperative to utilize diverse datasets in AI training to avoid biased educational outcomes. Educational stakeholders, including authorities, teachers, and students, must collaboratively engage to cultivate a more equitable and unbiased educational framework.

# 5 Discussion and conclusion

In summary, the integration of GAI into education has demonstrated transformative potential, particularly in the realm of personalized learning. Our review has revealed that GAI can effectively assist in generating individualized learning strategies, designing personalized learning paths, and enhancing teaching and learning through adaptive and context-aware methodologies. By supporting both educators and learners, GAI facilitates planning, implementation, and assessment, while also fostering complementary teacher-student relationships. Moreover, GAI aids learners by providing direct solutions, instructional cues, and heuristic dialogues, ultimately enabling a more engaging and responsive learning environment. Moving forward, educators and learners are encouraged to explore innovative ways of applying GAI in educational contexts, ensuring that its unique capabilities are harnessed to maximize personalized learning benefits. Addressing challenges in implementation and refining strategies will be key in unlocking the full potential of GAI in education.

## Acknowledgments

This work was partially supported by the Special Foundation for Interdisciplinary Talent Training in "AI Empowered Psychology / Education" of the School of Computer Science and Technology, East China Normal University, under the Grant 2024JCRC-03, and the Doctoral Research and Innovation Foundation of the School of Computer Science and Technology, East China Normal University, under the Grant 2023KYCX-03.

## References

Abdelghani, R., Wang, Y.-H., Yuan, X., Wang, T., Lucas, P., Sauzéon, H., & Oudeyer, P.-Y. (2023). Gpt-3-driven pedagogical agents to train children's curious question-asking skills. *International Journal of Artificial Intelligence in Education*, 1–36.

Afzal, S., Dhamecha, T., Mukhi, N., Sindhgatta Rajan, R., Marvaniya, S., Ventura, M., & Yarbro, J. (2019). Development and deployment of a large-scale dialog-based intelligent tutoring system. *Proceedings of the 2019 Conference of the North American Chapter of the Association for Computational Linguistics: Human Language Technologies, Volume 2 (Industry Papers)*, 114–121.

Ala-Mutka, K. M. (2005). A survey of automated assessment approaches for programming assignments. *Computer Science Education*, *15*(2), 83–102.

Al-Hossami, E., Bunescu, R., Teehan, R., Powell, L., Mahajan, K., & Dorodchi, M. (2023). Socratic questioning of novice debuggers: A benchmark dataset and preliminary evaluations. *Proceedings of the 18th Workshop on Innovative Use of NLP for Building Educational Applications (BEA 2023)*, 709–726.

Anderson, J. R., & Skwarecki, E. (1986). The automated tutoring of introductory computer programming. *Communications of the ACM*, *29*(9), 842–849.

Anghileri, J. (2006). Scaffolding practices that enhance mathematics learning. *Journal of Mathematics Teacher Education*, *9*, 33–52.

Bai, J. Y., Zawacki-Richter, O., & Muskens, W. (2022). Developing strategic scenarios for artificial intelligence applications in higher education. *5th International Open and Distance Learning Conference Proceedings Book*, 47–70.

Baidoo-Anu, D., & Ansah, L. O. (2023). Education in the era of generative artificial intelligence (AI): Understanding the potential benefits of ChatGPT in promoting teaching and learning. *Journal of AI*, *7*(1), 52–62.

Boston, M. D., & Candela, A. G. (2018). The Instructional Quality Assessment as a tool for reflecting on instructional practice. *ZDM*, *50*, 427–444.

Brown, T., Mann, B., Ryder, N., Subbiah, M., Kaplan, J. D., Dhariwal, P., Neelakantan, A., Shyam, P., Sastry, G., … Askell, A. (2020). Language models are few-shot learners. *Advances in Neural Information Processing Systems*, *33*, 1877–1901.

Burleigh, A. H. (1973). *Education In A Free Society*. New York. Longman Publishing Group.

Celik, I., Dindar, M., Muukkonen, H., & Järvelä, S. (2022). The promises and challenges of artificial intelligence for teachers: A systematic review of research. *TechTrends*, *66*(4), 616–630.

Chan, C. K. Y., & Tsi, L. H. (2023). The AI Revolution in Education: Will AI Replace or Assist Teachers in Higher Education? *arXiv Preprint arXiv:2305.01185*.

Chen, M., Tworek, J., Jun, H., Yuan, Q., Pinto, H. P. de O., Kaplan, J., Edwards, H., Burda, Y., Joseph, N., … Brockman, G. (2021). Evaluating large language models trained on code. *arXiv Preprint arXiv:2107.03374*.

Dargham, J., Saeed, D., & Mcheick, H. (2012). *E-Learning at school level: Challenges and Benefits* [Master's Thesis]. St. John Fisher University.

Diwan, C., Srinivasa, S., Suri, G., Agarwal, S., & Ram, P. (2023). AI-based learning content generation and learning pathway augmentation to increase learner engagement. *Computers and Education: Artificial Intelligence*, *4*, 100110.

Duran, R., Zavgorodniaia, A., & Sorva, J. (2022). Cognitive load theory in computing education research: A review. *ACM Transactions on Computing Education (TOCE)*, *22*(4), 1–27.

Edwards, J., Ditton, J., Trninic, D., Swanson, H., Sullivan, S., & Mano, C. (2020). Syntax exercises in CS1. *Proceedings of the 2020 ACM Conference on International Computing Education Research*, 216–226.

Ericsson, K. A., Krampe, R. T., & Tesch-Römer, C. (1993). The role of deliberate practice in the acquisition of expert performance. *Psychological Review*, *100*(3), 363.

Fariani, R. I., Junus, K., & Santoso, H. B. (2023). A Systematic Literature Review on Personalised Learning in the Higher Education Context. *Technology, Knowledge and Learning*, *28*(2), 449–476.

Gilbert, B. (2015). *Online learning revealing the benefits and challenges* [Master's Thesis]. St. John Fisher University.

Guo, K., Zhong, Y., Li, D., & Chu, S. K. W. (2023). Effects of chatbot-assisted in-class debates on students' argumentation skills and task motivation. *Computers & Education*, 104862.

Holstein, K., Aleven, V., & Rummel, N. (2020). A conceptual framework for human–AI hybrid adaptivity in education. *Artificial Intelligence in Education: 21st International Conference, AIED 2020, Ifrane, Morocco, July 6–10, 2020, Proceedings, Part I 21*, 240–254.

Hu, H., Jiang, Y.-H., & Li, R. (2024). Finetuning Large Language Models to Automatically Classify Cognitive Skills in Mathematical Problems. *Conference Proceedings of the 28th Global Chinese Conference on Computers in Education (GCCCE 2024)*, 145–152. http://gccce2024.swu.edu.cn/GCCCE2024_gongzuofanglunwenji2024-06-23A.pdf#page=163

Ihantola, P., Ahoniemi, T., Karavirta, V., & Seppälä, O. (2010). Review of recent systems for automatic assessment of programming assignments. *Proceedings of the 10th Koli Calling International Conference on Computing Education Research*, 86–93.

Jeon, J., & Lee, S. (2023). Large language models in education: A focus on the complementary relationship between human teachers and ChatGPT. *Education and Information Technologies*, 1–20.

Jiang, Y.-H. (2024). Multi-Agent System for Math Learning: Contextualized Mathematics Multiple-Choice Question Generation with Agentic Workflow. *2nd Global Summit on Artificial Intelligence*, 22. https://artificial-intelligence.hspioa.org/

Jiang, Y.-H., Li, R., Wei, Y., Jia, R., Shao, X., Hu, H., & Jiang, B. (2024). Synchronizing Verbal Responses and Board Writing for Multimodal Math Instruction with LLMs. *The 4th Workshop on Mathematical Reasoning and AI at NeurIPS'24*, 46–59. https://openreview.net/forum?id=esbIrV8N12

Jiang, Y.-H., Li, R., Zhou, Y., Qi, C., Hu, H., Wei, Y., Jiang, B., & Wu, Y. (2024). AI Agent for Education: Von Neumann Multi-Agent System Framework. *Conference Proceedings of the 28th Global Chinese Conference on Computers in Education (GCCCE 2024)*, 77–84. https://doi.org/10.48550/arXiv.2501.00083

Jiang, Y.-H., Shi, J., Tu, Y., Zhou, Y., Zhang, W., & Wei, Y. (2024). For Learners: AI Agent is All You Need. In Y. Wei, C. Qi, Y.-H. Jiang, & L. Dai (Eds.), *Enhancing Educational Practices: Strategies for Assessing and Improving Learning Outcomes* (pp. 21–46). Nova Science Publishers. https://doi.org/10.52305/RUIG5131

Jin, S.-H., Im, K., Yoo, M., Roll, I., & Seo, K. (2023). Supporting students' self-regulated learning in online learning using artificial intelligence applications. *International Journal of Educational Technology in Higher Education*, *20*(1), 1–21.

Kasneci, E., Seßler, K., Küchemann, S., Bannert, M., Dementieva, D., Fischer, F., Gasser, U., Groh, G., Günnemann, S., … Hüllermeier, E. (2023). ChatGPT for good? On opportunities and challenges of large language models for education. *Learning and Individual Differences*, *103*, 102274.

Kim, N. J., Belland, B. R., & Walker, A. E. (2018). Effectiveness of computer-based scaffolding in the context of problem-based learning for STEM education: Bayesian meta-analysis. *Educational Psychology Review*, *30*, 397–429.

Kochmar, E., Vu, D. D., Belfer, R., Gupta, V., Serban, I. V., & Pineau, J. (2022). Automated data-driven generation of personalized pedagogical interventions in intelligent tutoring systems. *International Journal of Artificial Intelligence in Education*, *32*(2), 323–349.

Koedinger, K. R., Brunskill, E., Baker, R. Sj., McLaughlin, E. A., & Stamper, J. (2013). New potentials for data-driven intelligent tutoring system development and optimization. *AI Magazine*, *34*(3), 27–41.

Li, R., Jiang, Y.-H., Wang, Y., Hu, H., & Jiang, B. (2024). A Large Language Model-Enabled Solution for the Automatic Generation of Situated Multiple-Choice Math Questions. *Conference Proceedings of the 28th Global Chinese Conference on Computers in Education (GCCCE 2024)*, 130–136. http://gccce2024.swu.edu.cn/GCCCE2024_gongzuofanglunwenji2024-06-23A.pdf#page=148

Li, R., Wang, Y., Zheng, C., Jiang, Y.-H., & Jiang, B. (2024). Generating Contextualized Mathematics Multiple-Choice Questions Utilizing Large Language Models. In A. M. Olney, I.-A. Chounta, Z. Liu, O. C. Santos, & I. I. Bittencourt (Eds.), *Artificial Intelligence in Education. Posters and Late Breaking Results, Workshops and Tutorials, Industry and Innovation Tracks, Practitioners, Doctoral Consortium and Blue Sky* (pp. 494–501). Springer Nature Switzerland. https://doi.org/10.1007/978-3-031-64315-6_48

Li, X., Guo, S., Wu, J., & Zheng, C. (2025). An interpretable polytomous cognitive diagnosis framework for predicting examinee performance. *Information Processing & Management*, *62*(1), 103913. https://doi.org/10.1016/j.ipm.2024.103913

Lister, R., Simon, B., Thompson, E., Whalley, J. L., & Prasad, C. (2006). Not seeing the forest for the trees: Novice programmers and the SOLO taxonomy. *ACM SIGCSE Bulletin*, *38*(3), 118–122.

Lu, X., Fan, S., Houghton, J., Wang, L., & Wang, X. (2023). Readingquizmaker: A human-nlp collaborative system that supports instructors to design high-quality reading quiz questions. *Proceedings of the 2023 CHI Conference on Human Factors in Computing Systems*, 1–18.

Markel, J. M., Opferman, S. G., Landay, J. A., & Piech, C. (2023). GPTeach: Interactive TA training with GPT-based students. *Proceedings of the Tenth Acm Conference on Learning@ Scale*, 226–236.

Marsh, T., Xuejin, C., Nickole, L. Z., Osterweil, S., Klopfer, E., & Haas, J. (2011). Fun and learning: The power of narrative. *Proceedings of the 6th International Conference on Foundations of Digital Games*, 23–29.

Mayer, R. E. (1987). The elusive search for teachable aspects of problem solving. *Historical Foundations of Educational Psychology*, 327–347.

Mayer, R. E. (1988). Learning strategies: An overview. *Learning and Study Strategies*, 11–22.

Mizumoto, A., & Eguchi, M. (2023). Exploring the potential of using an AI language model for automated essay scoring. *Research Methods in Applied Linguistics*, *2*(2), 100050.

Molenaar, I. (2022). Towards hybrid human-AI learning technologies. *European Journal of Education*, *57*(4), 632–645.

Nabizadeh, A. H., Leal, J. P., Rafsanjani, H. N., & Shah, R. R. (2020). Learning path personalization and recommendation methods: A survey of the state-of-the-art. *Expert Systems with Applications*, *159*, 113596.

Nabizadeh, A. H., Mário Jorge, A., & Paulo Leal, J. (2017). Rutico: Recommending successful learning paths under time constraints. *Adjunct Publication of the 25th Conference on User Modeling, Adaptation and Personalization*, 153–158.

Nikolic, S., Daniel, S., Haque, R., Belkina, M., Hassan, G. M., Grundy, S., Lyden, S., Neal, P., & Sandison, C. (2023). ChatGPT versus engineering education assessment: A multidisciplinary and multi-institutional benchmarking and analysis of this generative artificial intelligence tool to investigate assessment integrity. *European Journal of Engineering Education*, 1–56.

Nurmayanti, N., & Suryadi, S. (2023). The Effectiveness Of Using Quillbot In Improving Writing For Students Of English Education Study Program. *Jurnal Teknologi Pendidikan: Jurnal Penelitian Dan Pengembangan Pembelajaran*, *8*(1), 32–40.

Paiva, J. C., Leal, J. P., & Figueira, Á. (2022). Automated assessment in computer science education: A state-of-the-art review. *ACM Transactions on Computing Education (TOCE)*, *22*(3), 1–40.

Pankiewicz, M., & Baker, R. S. (2023). Large Language Models (GPT) for automating feedback on programming assignments. *arXiv Preprint arXiv:2307.00150*.

Patel, N., Nagpal, P., Shah, T., Sharma, A., Malvi, S., & Lomas, D. (2023). Improving mathematics assessment readability: Do large language models help? *Journal of Computer Assisted Learning*, *39*(3), 804–822.

Phung, T., Cambronero, J., Gulwani, S., Kohn, T., Majumdar, R., Singla, A., & Soares, G. (2023). Generating High-Precision Feedback for Programming Syntax Errors using Large Language Models. *arXiv Preprint arXiv:2302.04662*.

Poesia, G., Polozov, O., Le, V., Tiwari, A., Soares, G., Meek, C., & Gulwani, S. (2022). Synchromesh: Reliable code generation from pre-trained language models. *arXiv Preprint arXiv:2201.11227*.

Qi, C., Jia, L., Wei, Y., Jiang, Y.-H., & Gu, X. (2024). IntelliChain: An Integrated Framework for Enhanced Socratic Method Dialogue with LLMs and Knowledge Graphs. *Conference Proceedings of the 28th Global Chinese Conference on Computers in Education (GCCCE 2024)*, 116–121. http://gccce2024.swu.edu.cn/GCCCE2024_gongzuofanglunwenji2024-06-23A.pdf#page=134

Quintana, C., Reiser, B. J., Davis, E. A., Krajcik, J., Fretz, E., Duncan, R. G., Kyza, E., Edelson, D., & Soloway, E. (2018). A scaffolding design framework for software to support science inquiry. In *Scaffolding* (pp. 337–386). Psychology Press.

Radford, A., Wu, J., Child, R., Luan, D., Amodei, D., … Sutskever, I. (2019a). Language models are unsupervised multitask learners. *OpenAI Blog*, *1*(8), 9.

Radford, A., Wu, J., Child, R., Luan, D., Amodei, D., Sutskever, I., & others. (2019b). Language models are unsupervised multitask learners. *OpenAI Blog*, *1*(8), 9.

Sarsa, S., Denny, P., Hellas, A., & Leinonen, J. (2022). Automatic Generation of Programming Exercises and Code Explanations with Large Language Models. *arXiv Preprint arXiv:2206.11861*.

Schroeder, K. T., Hubertz, M., Van Campenhout, R., & Johnson, B. G. (2022). Teaching and Learning with AI-Generated Courseware: Lessons from the Classroom. *Online Learning*, *26*(3), 73–87.

Sridhar, P., Doyle, A., Agarwal, A., Bogart, C., Savelka, J., & Sakr, M. (2023). Harnessing llms in curricular design: Using gpt-4 to support authoring of learning objectives. *arXiv Preprint arXiv:2306.17459*.

Sternberg, R. J. (1982). *How and how much can intelligence be increased*. Ablex Publishing Corporation.

Tan, C. (2017). Confucianism and education. In *Oxford research encyclopedia of education*.

Thakkar, S. R., & Joshi, H. D. (2015). E-learning systems: A review. *2015 IEEE Seventh International Conference on Technology for Education (T4E)*, 37–40.

Tlili, A., Shehata, B., Adarkwah, M. A., Bozkurt, A., Hickey, D. T., Huang, R., & Agyemang, B. (2023). What if the devil is my guardian angel: ChatGPT as a case study of using chatbots in education. *Smart Learning Environments*, *10*(1), 15.

van den Berg, G., & du Plessis, E. (2023). ChatGPT and generative AI: Possibilities for its contribution to lesson planning, critical thinking and openness in teacher education. *Education Sciences*, *13*(10), 998.

Vygotsky, L. S., & Cole, M. (1978). *Mind in society: Development of higher psychological processes*. Harvard university press.

Wang, W., Tu, Y., Song, L., Zheng, J., & Wang, T. (2021). An Adaptive Design for Item Parameter Online Estimation and Q-Matrix Online Calibration in CD-CAT. *Frontiers in Psychology*, *12*, 710497. https://doi.org/10.3389/fpsyg.2021.710497

Wang, W., Zheng, J., Song, L., Tu, Y., & Gao, P. (2021). Test Assembly for Cognitive Diagnosis Using Mixed-Integer Linear Programming. *Frontiers in Psychology*, *12*. https://doi.org/10.3389/fpsyg.2021.623077

Wardat, Y., Tashtoush, M. A., AlAli, R., & Jarrah, A. M. (2023). ChatGPT: A revolutionary tool for teaching and learning mathematics. *Eurasia Journal of Mathematics, Science and Technology Education*, *19*(7), em2286.

Wei, Y., Zhou, Y., Jiang, Y.-H., & Jiang, B. (2024). Enhancing Explainability of Knowledge Learning Paths: Causal Knowledge Networks. *Joint Proceedings of the Human-Centric eXplainable AI in Education and the Leveraging Large Language Models for Next Generation Educational Technologies Workshops (HEXED-L3MNGET 2024) Co-Located with 17th International Conference on Educational Data Mining (EDM 2024)*, *3840*, 9–17. https://doi.org/10.48550/arXiv.2406.17518

Wood, T. (1994). Patterns of interaction and the culture of mathematics classrooms. *Cultural Perspectives on the Mathematics Classroom*, 149–168.

Zhou, Y., Zhang, M., Jiang, Y.-H., Liu, N., & Jiang, B. (2024). A Study on Educational Data Analysis and Personalized Feedback Report Generation Based on Tags and ChatGPT. *Conference Proceedings of the 28th Global Chinese Conference on Computers in Education (GCCCE 2024)*, 108–115. http://gccce2024.swu.edu.cn/GCCCE2024_gongzuofanglunwenji2024-06-23A.pdf#page=126